\pdfoutput=1

\documentclass[11pt]{article}

\usepackage[]{EMNLP2023}

\usepackage{times}
\usepackage{latexsym}

\usepackage[T1]{fontenc}

\usepackage[utf8]{inputenc}

\usepackage{microtype}

\usepackage{inconsolata}

\usepackage{tabularray}
\usepackage{tabularx}
\title{Large Language Models for Propaganda Detection}

\author{Author 1 \and ... \and Author n \\
        Address line \\ ... \\ Address line}

\author{Kilian Sprenkamp \\
  University of Zurich \\
  \texttt{kilian.sprenkamp@uzh.ch} \\\And
  Daniel Gordon Jones \\
  University of Zurich  \\
  \texttt{danielgordon.jones@uzh.ch} \\\And
  Liudmila Zavolokina \\
  University of Zurich  \\
  \texttt{zavolokina@ifi.uzh.ch}} 

\begin{document}
\maketitle
\begin{abstract}
The prevalence of propaganda in our digital society poses a challenge to societal harmony and the dissemination of truth. Detecting propaganda through NLP in text is challenging due to subtle manipulation techniques and contextual dependencies. To address this issue, we investigate the effectiveness of modern Large Language Models (LLMs) such as GPT-3 and GPT-4 for propaganda detection. We conduct experiments using the SemEval-2020 task 11 dataset, which features news articles labeled with 14 propaganda techniques as a multi-label classification problem. Five variations of GPT-3 and GPT-4 are employed, incorporating various prompt engineering and fine-tuning strategies across the different models. We evaluate the models' performance by assessing metrics such as $F1$ score, $Precision$, and $Recall$, comparing the results with the current state-of-the-art approach using RoBERTa. Our findings demonstrate that GPT-4 achieves comparable results to the current state-of-the-art. Further, this study analyzes the potential and challenges of LLMs in complex tasks like propaganda detection.
\end{abstract}

\section{Introduction}
Propaganda has long been a pernicious tactic of authoritarian regimes, used to manipulate public opinion, legitimize political power, and stifle dissent \cite{stanley2015propaganda}. Propaganda can create a distorted and often misleading picture of reality that reinforces the authority of the ruling elite and undermines the ability of the population to challenge it. This in turn leads to increased polarization and division, reduced critical thinking skills, and the dehumanization of others. As such, the detection and exposure of propaganda has emerged as a critical task in the digital information ecosystem (e.g. \citet{da2020prta}).

Recently Large Language Models (LLMs) such as GPT \cite{Brown2020Language, openai2023gpt4} have shown remarkable capabilities in various NLP tasks. Due to self-supervised pre-training on large amounts of text, LLMs achieve state-of-the-art results in zero and few-shot scenarios \cite{Brown2020Language}. However, the application of LLMs to non-standard tasks, such as propaganda detection framed as a multi-label classification problem, remains underexplored. Furthermore, the unique potential and challenges of LLMs in such complex tasks have not been thoroughly investigated. Our study aims to fill this knowledge gap by providing structured experiments applying prompt-based learning \cite{liu2023pre} using multiple versions of GPT-3 and GPT-4.
\section{Background}
\subsection{Propaganda Detection Approaches}
Propaganda is the intentional influencing of someone's opinion using various rhetorical and psychological techniques \cite{DaSanMartino2020Prta}. Propaganda uses techniques such as loaded language (using words or phrases with strong emotional connotations to influence an audience's opinion) or flag waving (associating oneself or one's cause with patriotism or a national symbol to gain support) to manipulate perceptions and shape public attitudes \cite{da2019fine}. 
Investigative initiatives such as PolitiFact \footnote{\tiny{https://www.politifact.com/}} or Bellingcat \footnote{\tiny{https://www.bellingcat.com/}} counter propaganda through manual fact-checking and play a crucial role in exposing propaganda techniques to the public. However, these initiatives often rely on labor-intensive analysis and volunteer research, which is not scalable.

The ongoing discourse on automated propaganda detection revolves around framing the task as either supervised token or multi-label classification problems \cite{martino2020semeval,alam2022overview,al2019justdeep,da2019fine}.The given labeled datasets focus predominantly on the detection of propaganda within the English language \cite{al2019justdeep, martino2020semeval}. Obtaining labeled data for this purpose is challenging, time and cost consuming, and a largely subjective task \cite{Ahmed2021Detecting}. The \textit{SemEval-2020 task 11: Detection of propaganda techniques in news articles} \cite{martino2020semeval}, currently stands as the benchmark supervised dataset for propaganda detection. It defines propaganda through 14 techniques outlined in Table \ref{Table: Propaganda Techniques} within the Appendix, and includes labeled articles specifically regarding these techniques. For the task of multi-label propaganda detection, various models have been developed including shallow learning models (e.g. Random Forest or SVM) \cite{ermurachi2020uaic1860}, CNN and LSTM models \cite{abedalla2019closer}, as well as BERT-based models \cite{abdullah2022detecting, da2020prta}. For the \textit{SemEval-2020 task 11} dataset, the current state-of-the-art model is created by \citet{abdullah2022detecting}, which fine-tuned RoBERTa \cite{liu2019roberta} with inputs being truncated to 120 tokens using the Roberta Tokenizer for the given task. However, to the best of our knowledge, the usage of LLMs for propaganda detection has not yet been tested.

\subsection{Large Language Models for Text Classification}
The advent of LLMs has brought about significant advancements in the field of NLP. LLMs leverage the transformer architecture \cite{vaswani2017attention} and are trained on vast amounts of text (e.g. \citet{gao2020pile}) using self-supervision \cite{yarowsky1995unsupervised,liuselfsuper}. Having learned rich linguistic features, LLMs can be used for various NLP tasks within zero and few-shot learning settings \cite{radford2019language,Brown2020Language}, reducing the need for training data to achieve state-of-the-art results.

The increasing proficiency of LLMs has sparked a rising interest in the area of prompt-based learning \cite{liu2023pre}. Unlike the standard supervised learning approach, which models $P(y|x; \Theta)$, prompt-based learning with LLMs instead models $P(x; \Theta)$, using this probability to predict $y$ \cite{liu2023pre}. While transfer learning with LLMs within a standard supervised learning framework can yield impressive results; this is often achieved by adapting the last layer of the neural network according to the task (e.g. \citet{kant2018practical, raffel2020exploring}). However, our focus will be on exploring the potential of prompt-based learning approaches. 

Prompt engineering, a key component of prompt-based learning, is a systematic process of creating prompts based on a variety of established patterns to boost the performance of a model \cite{liu2022design}. The two most prevalent strategies for prompting LLMs are the `few-shot' and `chain of thought' pattern \cite{wei2022chain}. In the `few-shot' pattern, the prompt features a series of examples regarding $x$ and $y$. These examples serve as a roadmap, helping to steer the LLM toward the intended task, and presenting an ideal template for input and output formatting \cite{wei2022chain}. The `chain of thought' pattern, however, adopts a more prescriptive strategy, using a series of clear instructions to guide the LLM though the process. This method stimulates the LLM to reflect upon its output, often producing more refined and accurate results \cite{wei2022chain}. Moreover, the `chain of thought' pattern can be used as a tool for explainability, enabling the model to generate an explanation of the given output \cite{liang2021explainable}.

Prompt-based approaches for text classification have been applied recently within various fields such as medicine \cite{sivarajkumar2022healthprompt}, law \cite{trautmann2022legal}, or human resource management \cite{clavie2023large}. Utilizing LLMs via prompt engineering for text classification has the benefit of being highly adaptive towards a given task without the need of transfer learning \cite{puri2019zero}. However, as the primary functions of LLMs are the analysis and generation of text \cite{Brown2020Language}, there are numerous factors that determine the model performance. The results of LLMs are not inherently deterministic and standardized. For text classification, this could mean that the output does not fall in the predefined set of categories or is reproducible over multiple inference calls. Furthermore, LLMs have the chance of output repetitions, hallucinations, and unanticipated terminations in output generation due to vanishing gradients \cite{lee2023mathematical} or the underlying training data \cite{ji2023survey}. A possible solution is proposed by reinforcement learning with human feedback (RLHF) \cite{christiano2017deep}, which enables better alignment with the user's intent \cite{openai2023gpt4}.
\newpage

\section{Methodology}
To analyze the potentials and challenges of LLMs for the task of multi-label text classification in propaganda detection we use the \textit{SemEval-2020 task 11: Detection of propaganda techniques in news articles} dataset \cite{martino2020semeval}. The dataset is split into 371 training (we kept 20\% for validation purposes) and 75 articles for testing. We employ five variations of GPT-3 fine-tuned \cite{Brown2020Language} and GPT-4 out-of-the box \cite{openai2023gpt4} to detect propaganda techniques within the news articles. 

For GPT-4, we employ two distinct types of prompts. In one approach, the model is solely tasked with outputting the labels for a given article, referred to as the `base' prompt. In a complementary approach, we have a `chain of thought' prompt that instructs the model to engage in a reasoning process about the predicted labels. For both types of prompts we apply the `few-shot' pattern, giving a single example for each propaganda technique within the prompt (Table \ref{Table: Propaganda Techniques}).

Additionally, we fine-tuned GPT-3 Davinci with the given dataset and employ the  `base' and `chain of thought' prompts at the inference stage. It's worth noting that the OpenAI API \cite{OpenAIAPI} suggests that training only requires defining the input and output, without an instructional prompt. To investigate this proposition, we fine-tuned another model without any instruction given at training and inference. Furthermore, for the GPT-3 models, we split the input data into multiple chunks due to the maximum token limit of $2048$, compared to $8192$ for GPT-4. To ensure reproducible outputs, we set the $temperature$ parameter to $0$, such that the output becomes more deterministic.
We benchmark our five models against the current state-of-the-art approach by \citet{abdullah2022detecting}, which investigated the capabilities of several transformer \cite{vaswani2017attention} based models, with RoBERTa \cite{liu2019roberta} scoring the highest $F_1$ score. We assess the performance of all models using the micro average $F_1$ score, similar to \citet{abdullah2022detecting}, we additionally analyze the micro average $Precision$ and $Recall$ to better understand the working of the employed models. Lastly, we compare $F_1$ scores per label (i.e. the propaganda technique) of each model. All experiments can be replicated through the following GitHub repository\footnote{\tiny{https://github.com/sprenkamp/LLM\_propaganda\_detection}}.
\section{Experiments}
The results from our proposed experiments are depicted in Table~\ref{Table: Performance}. It is discernible that both versions of GPT-4 significantly exceed the performance of fine-tuned GPT-3 models, with the GPT-4 `base' model achieving the highest $F1$ score among the models we developed. 

\begin{table}[h]
\centering
\scriptsize
\begin{tabular}{|l|l|l|l|}
\hline
\textbf{Model} & \textbf{Precision} & \textbf{Recall} & \textbf{F1 Score}\\
\hline
GPT-4 base & 52.86\% & 64.52\% & 58.11\% \\
GPT-4 chain of thought & 56.86\% & 57.82\% & 57.34\% \\
GPT-3 base & 44.35\% & 44.00\% & 44.18\% \\
GPT-3 chain of thought & 48.62\% & 28.16\% & 35.66\% \\
GPT-3 no instruction & 47.54\% & 32.48\% & 38.59\% \\
Baseline \cite{abdullah2022detecting} & NA & NA & 63.40\%\\
\hline
\end{tabular}
\caption{Performance of different models. GPT-3 was fine-tuned, while GPT-4 was employed out-of-the-box}
\label{Table: Performance}
\end{table}

\textbf{GPT-3.} Further analysis revealed a tendency of the fine-tuned GPT-3 models to overfit to the training data, which led to a propensity to predict only those labels that were over-represented in the training set (for instance, Loaded\textunderscore Language). Additional scrutiny of the GPT-3 models' output uncovered issues such as repetitions, hallucinations, and unanticipated terminations in output generation, even when the maximum token limit had not been met; indicating that the models can suffer from vanishing gradients. This issue is particularly conspicuous when comparing the output from GPT-3 and GPT-4 in the `chain of thought' models. While GPT-4 generated plausible reasoning for each classified label, GPT-3 was unable to do so.

\textbf{GPT-4.} Comparing the two GPT-4 models, we noticed that the `chain of thought' prompt led to a higher $Precision$, suggesting that the model is more discerning and tends to classify a label only when it can provide sound reasoning for it. In contrast, the GPT-4 `base' model displayed a higher $Recall$, indicating a more liberal approach to prediction, resulting in a better capture of positive instances, albeit at the cost of a higher number of false positives. These disparities are further illuminated when comparing the $F1$ score for each label, as shown in Table~\ref{Table: F1 Scores per Technique} in the Appendix. For instance, the GPT-4 `chain of thought' model scored an $F1$ Score of $0 \%$ for five propaganda techniques, implying an inability to reason about these concepts.

\textbf{Baseline Comparison.} While the $F1$ scores of both GPT-4 models are similar, they do not surpass the current state-of-the-art method based on RoBERTa \cite{abdullah2022detecting}. Nevertheless, when we juxtaposed the $F1$ Scores per label of the GPT-4 `base' model with the method of \citet{abdullah2022detecting}, our model demonstrated superior performance for seven out of 14 propaganda techniques.

\section{Discussion}
Based on the experiments, we identified potentials and challenges of LLMs for propaganda detection in multi-label classification problems.

\textbf{Potentials.} Our results show that prompt-based learning, especially GPT-4, can yield results comparable to traditional supervised-learning methods. This is advantageous as it significantly reduces the explicit need for labeled training data and training resources. Further, by setting the $temperature$ parameter to $0$, we reduce the fear of non-deterministic results. With the ongoing development of the OpenAI API, reproducibility will even increase as the output can be predefined within functions \cite{openai2023function}. Moreover, our `chain of thought' approach elucidates the reasoning behind flagging a particular text as propaganda. As an extension of \citet{abdullah2022detecting}, we enable the model to classify texts and generate its own interpretations.However, potential dangers arise from this reasoning, as it may result from hallucinations \cite{lee2023mathematical,ji2023survey} caused by the LLM, posing a risk of over-reliance by the end user.

Moreover, GPT-4 being superior in the understanding of linguistic features over GPT-3 proves beneficial for the subjective task of propaganda detection, where general knowledge outweighs the need for fine-tuning. We wish to highlight RLHFs crucial role in driving the performance difference between the GPT-3 and GPT-4 models, as it was only applied to GPT-4 \cite{Brown2020Language, openai2023gpt4}. 

Regarding the usage of GPT-4, while the `base' model performs slightly better on $Recall$ and the $F1$ score, the `chain of thought' model boasts higher $Precision$. This highlights that the model's prediction tendency can be adjusted by refining the prompt instruction, depending on the importance of $Precision$ or $Recall$ in the specific task.

\textbf{Challenges.}
Our research indicates that GPT-3 models suffer from vanishing gradients, leading to repetitions, hallucinations, and abrupt stops within the output, making them unsuitable for propaganda detection \cite{lee2023mathematical,ji2023survey}. Additionally, with just 371 articles available for training, the data seems insufficient to learn the complex nuances of different propaganda techniques and tends to overfit the data.

Additionally, we see the maximum token length as a potentially limiting factor for successful propaganda detection within newspaper articles. Given that our prompt instruction in its `base' form is $\sim$ 740 token long, this leaves just 1308 tokens in the case of GPT-3 and 7452 in the case of GPT-4, for the article that we want to analyse and the completion returned by the model. However, the maximum token length is much higher compared to \citet{abdullah2022detecting}, with a maximum token length of 140. For multi-label classification problems the number of tokens used within prompt instruction will scale approximately linearly as each label would need a description and possibly an example, depending on the underlying task.

Furthermore, the performance of GPT-4 varies significantly across different propaganda techniques compared to the baseline model \cite{abdullah2022detecting} (Table \ref{Table: F1 Scores per Technique}). Some propaganda techniques (e.g., Bandwagon,Reductio\_ad\_hitlerum) are likely not well-represented in the model's training data, which is not disclosed in \citet{openai2023gpt4}. To address this issue, three strategies can be considered. Firstly, fine-tuning GPT-4 with our propaganda dataset, which may be feasible in the near future. Secondly, adapting the given prompt that currently relies solely on the definitions and examples of propaganda techniques by \cite{da2019fine, abdullah2022detecting}. Thirdly, considering the subjective nature of propaganda classification, we could modify the classification schema to enhance the LLM's comprehension. This proposition has far-reaching implications for task formulation in NLP.
\section{Future Work}
We presented a first analysis of how LLMs can be used for propaganda detection. Our contribution is two-fold. First, we show that LLMs, especially GPT-4, achieve comparable results to the current state-of-the-art in the task of propaganda detection. Second, we derive a preliminary analysis of potentials and challenges of LLMs for multi-label classification problems. Our aim is to expand on this analysis within future publications with the goal of creating an LLM-based tool paralleling the approach of \citet{da2020prta}, yet uniquely integrating the use of LLMs, which can be used by the wide public to flag and explain propaganda in text.

\section*{Limitations}
Our manuscript has several limitations.

First, we applied propaganda detection exclusively to news articles in the English language. We identified two major contributors that determine the applicability of the method to other languages: language morphology and the representation of each language within the training data of GPT-3 \cite{Brown2020Language} and GPT-4 \cite{openai2023gpt4}.

Second, we did not use any open-source models like LLaMA \cite{touvron2023llama}, which would enable more control over the training and inference process.

Third, we did not use the function call options within OpenAI API \cite{openai2023function}, we aim to do so in further model iterations.

Last, while our experiments have not been limited by any computational budget, i.e. cost of the OpeanAI API \citet{OpenAIAPI} we see this as potential limiting factor for future large scale resource.

\section*{Ethics Statement}
Regarding our manuscript, we see a general positive broader impact as it forms the basis for more effective detection of propaganda beneficial for society. However, we would like to state that limitations like misclassifications or hallucinations can impact the performance of LLMs for propaganda detection and have a negative impact on end users. Moreover, we fear that the application of automated propaganda detection can lead to an over-reliance on the end user side.

\bibliography{anthology,custom}
\bibliographystyle{acl_natbib}

\appendix
\section*{Appendix}
\label{sec:appendix}

\begin{table*}[t]
\centering
\small
\begin{tabular}{|l|p{5cm}|p{5cm}|}
\hline
\textbf{Propaganda Technique} & \textbf{Definition} & \textbf{Example} \\
\hline
Appeal\_to\_Authority & Supposes that a claim is true because a valid authority or expert on the issue supports it & "The World Health Organisation stated, the new medicine is the most effective treatment for the disease." \\ \hline
Appeal\_to\_fear-prejudice & Builds support for an idea by instilling anxiety and/or panic in the audience towards an alternative & "Stop those refugees; they are terrorists." \\ \hline
Bandwagon,Reductio\_ad\_hitlerum & Justify actions or ideas because everyone else is doing it, or reject them because it's favored by groups despised by the target audience & "Would you vote for Clinton as president? 57\% say yes." \\ \hline
Black-and-White\_Fallacy &  Gives two alternative options as the only possibilities, when actually more options exist & "You must be a Republican or Democrat" \\ \hline
Causal\_Oversimplification &  Assumes a single reason for an issue when there are multiple causes & "If France had not declared war on Germany, World War II would have never happened." \\ \hline
Doubt & Questioning the credibility of someone or something & "Is he ready to be the Mayor?" \\ \hline
Exaggeration,Minimisation & Either representing something in an excessive manner or making something seem less important than it actually is & "I was not fighting with her; we were just playing." \\ \hline
Flag-Waving & Playing on strong national feeling (or with respect to a group, e.g., race, gender, political preference) to justify or promote an action or idea & "Entering this war will make us have a better future in our country." \\ \hline
Loaded\_Language & Uses specific phrases and words that carry strong emotional impact to affect the audience & "A lone lawmaker"s childish shouting." \\ \hline
Name\_Calling,Labeling & Gives a label to the object of the propaganda campaign as either the audience hates or loves & "Bush the Lesser." \\ \hline
Repetition &  Repeats the message over and over in the article so that the audience will accept it & "Our great leader is the epitome of wisdom. Their decisions are always wise and just." \\ \hline
Slogans & A brief and striking phrase that contains labeling and stereotyping &  "Make America great again!" \\ \hline
Thought-terminating\_Cliches &  Words or phrases that discourage critical thought and useful discussion about a given topic & "It is what it is" \\ \hline
Whataboutism,Straw\_Men,Red\_Herring & Attempts to discredit an opponent's position by charging them with hypocrisy without directly disproving their argument & "They want to preserve the FBI's reputation." \\ 
\hline
\end{tabular}
\caption{Propaganda Techniques after \citet{da2019fine,abdullah2022detecting}}
\label{Table: Propaganda Techniques}
\end{table*}

\begin{table*}[]
\centering
\small
\begin{tabular}{|l|r|r|r|r|r|r|}
\hline
\textbf{Propaganda Technique} & \textbf{GPT-4 base} & \textbf{\begin{tabular}[c]{@{}r@{}}GPT-4\\ chain of \\ thought\end{tabular}} & \textbf{GPT-3 base} & \textbf{\begin{tabular}[c]{@{}r@{}}GPT-3 \\ chain of \\ thought\end{tabular}} & \textbf{\begin{tabular}[c]{@{}r@{}}GPT-3 \\ no \\instruction\end{tabular}} & \textbf{Baseline} \\ \hline
Appeal\_to\_Authority & 23.52\% & 19.05\% & 11.11\% & 16.67\% & 0.00\% & 47.36\%\\ 
Appeal\_to\_fear-prejudice & 52.50\% & 0.00\% & 8.70\% & 0.00\% & 0.00\% & 43.6\%\\
Bandwagon,Reductio\_ad\_hitlerum & 0.00\% & 0.00\% & 0.00\% & 0.00\% & 0.00\% & 4.87\% \\
Black-and-White\_Fallacy & 54.54\% & 0.00\% & 12.5\% & 0.00\% & 10.00\% & 24.09\% \\
Causal\_Oversimplification & 32.55\% & 50.00\% & 0.00\% & 0.00\% & 0.00\% & 19.44\% \\
Doubt & 52.63\% & 54.54\% & 16.98\% & 19.42\% & 16.22\% & 61.36\% \\
Exaggeration,Minimisation & 64.00\% & 64.00\% & 16.28\% & 8.22\% & 0.00\% & 33.03\%\\
Flag-Waving & 32.00\% & 0.00\% & 43.48\% & 0.00\% & 26.96\% & 61.49\%\\
Loaded\_Language & 92.75\% & 93.62\% & 73.51\% & 71.39\% & 70.02\% & 75.71\%\\
Name\_Calling,Labeling & 77.67\% & 74.29\% & 54.96\% & 31.32\% & 50.00\% & 67.49\%\\
Repetition & 56.67\% & 65.79\% & 31.88\% & 16.00\% & 17.78\% & 31.14\%\\
Slogans & 9.09\% & 10.00\% & 21.74\% & 21.74\% & 6.67\% & 54.90\%\\
Thought-terminating\_Cliches & 20.00\% & 0.00\% & 0.00\% & 0.00\% & 0.00\% & 25\%\\
Whataboutism,Straw\_Men,Red\_Herring & 14.81\% & 33.33\% & 0.00\% & 0.00\% & 0.00\% & 20.83\%\\
\hline
\end{tabular}
\caption{F1 Scores per Propaganda Technique. Note: The Baseline \citet{abdullah2022detecting} solely provide $F1 \, Scores$ for a test data set of which the labels are not publicly available, thus only a limited comparison is possible.}
\label{Table: F1 Scores per Technique}
\end{table*}

\end{document}